\documentclass[runningheads]{llncs}
\usepackage{graphicx}
\usepackage{epsfig}
\usepackage{graphicx}
\usepackage{multirow}
\usepackage{gensymb}
\usepackage{bm}
\usepackage{bbding}
\usepackage{booktabs}
\usepackage{xspace}

\usepackage{tikz}
\usepackage{comment}
\usepackage{amsmath,amssymb}
\usepackage{color}
\usepackage{cite}
\usepackage[section]{placeins}
\usepackage{url}

\usepackage[accsupp]{axessibility}

\usepackage[symbol]{footmisc}

\usepackage{array}
\newcolumntype{?}{!{\vrule width 1.3pt}}
\usepackage[pagebackref=true,breaklinks=true,letterpaper=true,colorlinks,bookmarks=false]{hyperref}
\usepackage{orcidlink}

\newcolumntype{M}[1]{>{\centering\arraybackslash}m{#1}}

\RequirePackage{makecell}

\begin{document}
\pagestyle{headings}
\mainmatter
\def\ECCVSubNumber{2558}

\title{DCL-Net: Deep Correspondence Learning Network for 6D Pose Estimation}

\author{
{Hongyang Li \inst{1}}\protect\footnotemark[1]\orcidlink{0000-0003-3231-0655} \and
Jiehong Lin \inst{1,2}\protect\footnotemark[1]\orcidlink{0000-0001-7495-6070} \and
Kui Jia\inst{1,3}\protect\footnotemark[2]\orcidlink{0000-0003-2661-5700}
}
\authorrunning{H. Li et al.}
\institute{South China University of Technology \and
DexForce Co. Ltd. \and 
Peng Cheng Laboratory \\
\email{\{eeli.hongyang,lin.jiehong\}@mail.scut.edu.cn, kuijia@scut.edu.cn}}
\maketitle
\footnotetext[1]{Equal contribution}
\footnotetext[2]{Corresponding author}

\begin{abstract}
  Establishment of point correspondence between camera and object coordinate systems is a promising way to solve 6D object poses. However, surrogate objectives of correspondence learning in 3D space are a step away from the true ones of object pose estimation, making the learning suboptimal for the end task.
  In this paper, we address this shortcoming by introducing a new method of \emph{Deep Correspondence Learning Network} for direct 6D object pose estimation, shortened as \emph{DCL-Net}. Specifically, DCL-Net employs dual newly proposed \emph{Feature Disengagement and Alignment (FDA) modules} to establish, in the feature space, partial-to-partial correspondence and complete-to-complete one for partial object observation and its complete CAD model, respectively, which result in aggregated pose and match feature pairs from two coordinate systems; these two FDA modules thus bring complementary advantages. 
  The match feature pairs are used to learn confidence scores for measuring the qualities of deep correspondence, while the pose feature pairs are weighted by confidence scores for direct object pose regression. A confidence-based pose refinement network is also proposed to further improve pose precision in an iterative manner. Extensive experiments show that DCL-Net outperforms existing methods on three benchmarking datasets, including YCB-Video, LineMOD, and Oclussion-LineMOD; ablation studies also confirm the efficacy of our novel designs. Our code is released publicly at \url{https://github.com/Gorilla-Lab-SCUT/DCL-Net}.

\end{abstract}

\section{Introduction}
\label{sec:Introduction}
6D object pose estimation is a fundamental task of 3D semantic analysis with many real-world applications, such as robotic grasping \cite{moped, wu2020grasp}, augmented reality \cite{handsAR}, and autonomous driving \cite{kitti,levinson2011towards,wang2019frustum,deng2022vista}. Non-linearity of the rotation space of $SO(3)$ makes it hard to handle this nontrivial task through direct pose regression from object observations \cite{discriminative, gradient, hausdorff,posecnn,densefusion,PRGCN,FSNet,dualPoseNet,SS-Conv,lin2022category}. Many of the data-driven methods \cite{hourglass, localaffine, orb,pvnet,learning6d, localrgbd, independent,pvn3d,NOCS, SPD} thus achieve the estimation by learning point correspondence between camera and object coordinate systems.

\begin{figure}[t]
  \centering
  \includegraphics[width=1.0\linewidth]{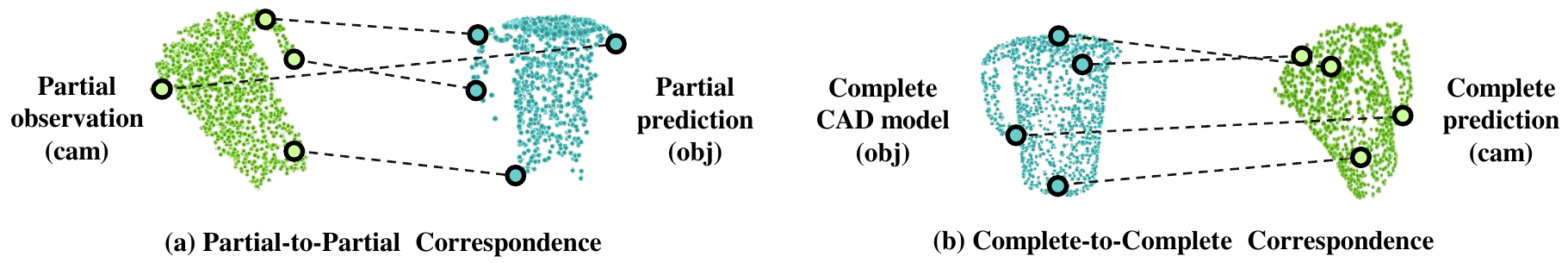}
    \caption{Illustrations of two kinds of point correspondence between camera coordinate system (cam) and object coordinate system (obj). Best view in the
    electronic version.}
    \label{fig:intro}
\end{figure}

Given a partial object observation in camera coordinate system along with its CAD model in object coordinate one, we show in Fig. \ref{fig:intro} two possible ways to build point correspondence: i) inferring the observed points in object coordinate system for partial-to-partial correspondence; ii) inferring the sampled points of CAD model in camera coordinate system for complete-to-complete correspondence. These two kinds of correspondence show different advantages. The partial-to-partial correspondence is of higher qualities than the complete-to-complete one due to the difficulty in shape completion, while the latter is more robust to figure out poses for objects with severe occlusions, which the former can hardly handle with. 

While these methods are promising by solving 6D poses from point correspondence (\eg, via a PnP algorithm), their surrogate correspondence objectives are a step away from the true ones of estimating 6D object poses, thus making their learnings suboptimal for the end task \cite{wang2021gdr}.
To this end, we present a novel method to realize the above two ways of correspondence establishment in the feature space via dual newly proposed \textit{Feature Disengagement and Alignment (FDA) modules}, and directly estimate object poses from feature pairs of two coordinate systems, which are weighted by confidence scores measuring the qualities of deep correspondence. We term our method as \textit{Deep Correspondence Learning Network}, shortened as \textit{DCL-Net}. Fig. \ref{fig:overview} gives the illustration.

For the partial object observation and its CAD model, DCL-Net firstly extracts their point-wise feature maps in parallel; then dual Feature Disengagement and Alignment (FDA) modules are designed to establish, in feature space, the partial-to-partial correspondence and the complete-to-complete one between camera and object coordinate systems. Specifically, each FDA module takes as inputs two point-wise feature maps, and disengages each feature map into individual pose and match ones; the match feature maps of two systems are then used to learn an attention map for building deep correspondence; finally, both pose and match feature maps are aligned and paired across systems based on the attention map, resulting in pose and match feature pairs, respectively. DCL-Net aggregates two sets of correspondence together, since they bring complementary advantages, by fusing the respective pose and match feature pairs of two FDA modules. The aggregated match feature pairs are used to learn confidence scores for measuring the qualities of deep correspondence, while the pose ones are weighted by the scores to directly regress object poses. A confidence-based pose refinement network is also proposed to further improve the results of DCL-Net in an iterative manner. Extensive experiments show that DCL-Net outperforms existing methods for 6D object pose estimation on three well-acknowledged datasets, including YCB-Video \cite{YCB}, LineMOD \cite{LineMod}, and Occlusion-LineMOD\cite{learning6d}; remarkably, on the more challenging Occlusion-LineMOD, our DCL-Net outperforms the state-of-the-art method \cite{FFB6D} with an improvement of $4.4\%$ on the metric of ADD(S), revealing the strength of DCL-Net on handling with occlusion. Ablation studies also confirm the efficacy of individual components of DCL-Net. Our technical contributions are summarized as follows:
\begin{itemize}
  \item We design a novel \textit{Feature Disengagement and Alignment (FDA) module} to establish deep correspondence between two point-wise feature maps from different coordinate systems; more specifically, FDA module disengages each feature map into individual pose and match ones, which are then aligned across systems to generate pose and match feature pairs, respectively, such that deep correspondence is established within the aligned feature pairs.
  \item We propose a new method of \textit{Deep Correspondence Learning Network} for direct regression of 6D object poses, termed as DCL-Net, which employs dual FDA modules to establish, in feature space, partial-to-partial correspondence and complete-to-complete one between camera and object coordinate systems, respectively; these two FDA modules bring complementary advantages.
  \item Match feature pairs of dual FDA modules are aggregated and used for learning of confidence scores to measure the qualities of correspondence, while pose feature pairs are weighted by the scores for estimation of 6D pose; a confidence-based pose refinement network is also proposed to iteratively improve pose precision.
\end{itemize}

\section{Related Work}
\label{sec:RelatedWork}

\subsubsection{6D Pose Estimation from RGB Data} This body of works can be broadly categorized into three types: i) holistic methods \cite{discriminative, gradient, hausdorff} for directly estimating object poses; ii) keypoint-based methods \cite{hourglass, localaffine, orb}, which establish 2D-3D correspondence via 2D keypoint detection, followed by a PnP/RANSAC algorithm to solve the poses; iii) dense correspondence methods \cite{learning6d, localrgbd, independent, pvnet}, which make dense pixel-wise predictions and vote for the final results.

Due to loss of geometry information, these methods are sensitive to lighting conditions and appearance textures, and thus inferior to the RGB-D methods.

\subsubsection{6D Pose Estimation from RGB-D Data} Depth maps provide rich geometry information complementary to appearance one from RGB images. Traditional methods \cite{learning6d, LineMod, rios2013discriminatively, tejani2014latent,wohlhart2015learning } solve object poses by extracting features from RGB-D data and performing correspondence grouping and hypothesis verification. Earlier deep methods, such as PoseCNN \cite{posecnn} and SSD-6D \cite{ssd6d}, learn coarse poses firstly from RGB images, and refine the poses on point clouds by using ICP \cite{ICP} or MCN \cite{MCN}. Recently, learning deep features of point clouds becomes an efficient way to improve pose precision, especially for methods \cite{densefusion, PRGCN} of direct regression, which make efforts to enhance pose embeddings from deep geometry features, due to the difficulty in the learning of rotations from a nonlinear space. Wang \textit{et al.} present DenseFusion \cite{densefusion}, which fuses local features of RGB images and point clouds in a point-wise manner, and thus explicitly reasons about appearance and geometry information to make the learning more discriminative; due to the incomplete and noisy shape information, Zhou \textit{et al.} propose PR-GCN \cite{PRGCN} to polish point clouds and enhance pose embeddings via Graph Convolutional Network. On the other hand, dense correspondence methods show the advantages of deep networks on building the point correspondence in Euclidean space; for example, He \textit{et al.} propose PVN3D \cite{pvn3d} to regress dense keypoints, and achieve remarkable results. While promising, these methods are usually trained with surrogate objectives instead of the true ones of estimating 6D poses, making the learning suboptimal for the end task.

Our proposed DCL-Net borrows the idea from dense correspondence methods by learning deep correspondence in feature space, and weights the feature correspondence based on confidence scores for direct estimation of object poses. Besides, the learned correspondence is also utilized by an iterative pose refinement network for precision improvement.

\section{Deep Correspondence Learning Network}
\label{sec:method}

\begin{figure}[t]
  \centering
  \includegraphics[width=1.0\linewidth]{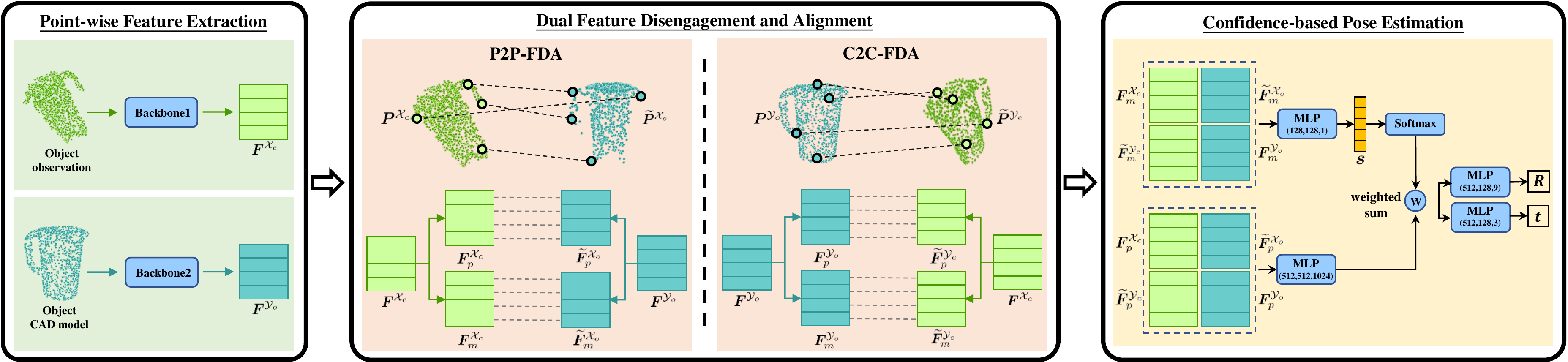}
    \caption{An illustration of DCL-Net. Given object observation and its CAD model, DCL-Net first extracts their point-wise features $\bm{F}^{\mathcal{X}_c}$ and $\bm{F}^{\mathcal{Y}_o}$, separately; then dual Feature Disengagement and Alignment (FDA) modules are employed to establish, in feature space, partial-to-partial correspondence and complete-to-complete one between camera and object coordinate systems, respectively, which result in aggregated pose and match feature pairs; the match feature pairs are used to learn confidence scores $\bm{s}$ for measuring the qualities of deep correspondence, while the pose ones are weighted by $\bm{s}$ for estimating 6D object pose $(\bm{R}, \bm{t})$. Best view in the electronic version.}
    \label{fig:overview}
\end{figure}

Given the partial object observation $\mathcal{X}_{c}$ in the camera coordinate system, along with the object CAD model $\mathcal{Y}_{o}$ in the object coordinate one, our goal is to estimate the 6D pose $(\bm{R}, \bm{t})$ between these two systems, where $\bm{R} \in SO(3)$ stands for a rotation, and $\bm{t} \in \mathbb{R}^3$ for a translation.

Fig. \ref{fig:overview} gives the illustration of our proposed \textit{Deep Correspondence Learning Network} (dubbed \textit{DCL-Net}). DCL-Net firstly extracts point-wise features of $\mathcal{X}_{c}$ and $\mathcal{Y}_{o}$ (cf. Sec. \ref{subsec:pointwise_feature_extraction}), then establishes correspondence in feature space via \textit{dual Feature Disengagement and Alignment modules} (cf. Sec. \ref{subsec:dual_feature_disengagement_and_alignment}), and finally regresses the object pose $(\bm{R}, \bm{t})$ with confidence scores based on the learned deep correspondence (cf. Sec. \ref{subsec:PoseEstimation}). The training objectives of DCL-Net are given in Sec. \ref{subsec:training_objectives}. A confidence-based pose refinement network is also introduced to iteratively improve pose precision (cf. Sec. \ref{subsec:confidence_based_pose_refinement}).

\subsection{Point-wise Feature Extraction}
\label{subsec:pointwise_feature_extraction}
We represent the inputs of the object observation $\mathcal{X}_{c}$ and its CAD model $\mathcal{Y}_{o}$ as $(\bm{I}^{\mathcal{X}_{c}}, \bm{P}^{\mathcal{X}_{c}})$ and $(\bm{I}^{\mathcal{Y}_{o}}, \bm{P}^{\mathcal{Y}_{o}})$ with $N_{\mathcal{X}}$ and $N_{\mathcal{Y}}$ sampled points, respectively, where $\bm{P}$ denotes a point set, and $\bm{I}$ denotes RGB values corresponding to points in $\bm{P}$.
As shown in Fig. \ref{fig:overview}, we use two parallel backbones to extract their point-wise features $\bm{F}^{\mathcal{X}_{c}}$ and $\bm{F}^{\mathcal{Y}_{o}}$, respectively.
Following \cite{SA-SSD}, both backbones are built based on 3D Sparse Convolutions \cite{SparseConv}, of which the volumetric features are then converted to point-level ones; more details about the architectures are given in the supplementary material. Note that for each object instance, $\bm{F}^{\mathcal{Y}_o}$ can be pre-computed during inference for efficiency.

\subsection{Dual Feature Disengagement and Alignment}
\label{subsec:dual_feature_disengagement_and_alignment}

The key to figure out the pose between the object observation and its CAD model lies in the establishment of correspondence. As pointed out in Sec. \ref{sec:Introduction}, there exist at least two ways to achieve this goal: i) learning the partial point set $\widetilde{\bm{P}}^{\mathcal{X}_{o}}$ in object system from complete $\bm{P}^{\mathcal{Y}_{o}}$ to pair with $\bm{P}^{\mathcal{X}_{c}}$, \eg, $(\bm{P}^{\mathcal{X}_{c}},\widetilde{\bm{P}}^{\mathcal{X}_{o}})$, for partial-to-partial correspondence; ii) inferring the complete point set $\widetilde{\bm{P}}^{\mathcal{Y}_{c}}$ in camera coordinate system from partial $\bm{P}^{\mathcal{X}_{c}}$ to pair with $\bm{P}^{\mathcal{Y}_{o}}$, \eg, $(\widetilde{\bm{P}}^{\mathcal{Y}_{c}}, \bm{P}^{\mathcal{Y}_{o}})$,  for complete-to-complete correspondence. 

In this paper, we propose to establish the correspondence in the deep feature space, from which \textit{pose feature pairs} along with \textit{match feature pairs} can be generated for the learning of object pose and confidence scores, respectively. Fig. \ref{fig:overview} gives illustrations of the correspondence in both 3D space and feature space.  
Specifically, we design a novel \textit{Feature Disengagement and Alignment (FDA) module} to learn the pose feature pairs, \eg, $(\bm{F}^{\mathcal{X}_{c}}_p, \widetilde{\bm{F}}^{\mathcal{X}_{o}}_p)$ and $(\widetilde{\bm{F}}^{\mathcal{Y}_c}_p, \bm{F}^{\mathcal{Y}_o}_p)$ \textit{w.r.t} the above $(\bm{P}^{\mathcal{X}_{c}},\widetilde{\bm{P}}^{\mathcal{X}_{o}})$ and $(\widetilde{\bm{P}}^{\mathcal{Y}_{c}},\bm{P}^{\mathcal{Y}_{o}})$, respectively, and the match feature pairs, \eg, $(\bm{F}^{\mathcal{X}_{c}}_m, \widetilde{\bm{F}}^{\mathcal{X}_{o}}_m)$ and $(\widetilde{\bm{F}}^{\mathcal{Y}_c}_m, \bm{F}^{\mathcal{Y}_o}_m)$, which can be formulated as follows:
\begin{equation}
  \bm{F}^{\mathcal{X}_{c}}_p, \bm{F}^{\mathcal{X}_{c}}_m, \widetilde{\bm{F}}^{\mathcal{X}_{o}}_p, \widetilde{\bm{F}}^{\mathcal{X}_{o}}_m, \widetilde{\bm{P}}^{\mathcal{X}_{o}} = \texttt{FDA}(\bm{F}^{\mathcal{X}_{c}}, \bm{F}^{\mathcal{Y}_{o}}),
\label{EqnPartialFDA}
\end{equation}
\begin{equation}
  \bm{F}^{\mathcal{Y}_o}_p, \bm{F}^{\mathcal{Y}_o}_m,\widetilde{\bm{F}}^{\mathcal{Y}_c}_p, \widetilde{\bm{F}}^{\mathcal{Y}_c}_m, \widetilde{\bm{P}}^{\mathcal{Y}_c} = \texttt{FDA}(\bm{F}^{\mathcal{Y}_o}, \bm{F}^{\mathcal{X}_c}).
  \label{EqnCompleteFDA}
\end{equation}
We term the partial-to-partial (\ref{EqnPartialFDA}) and complete-to-complete (\ref{EqnCompleteFDA}) FDA modules as P2P-FDA and C2C-FDA modules, respectively.

\begin{figure}[t]
  \centering
  \includegraphics[width=0.99\linewidth]{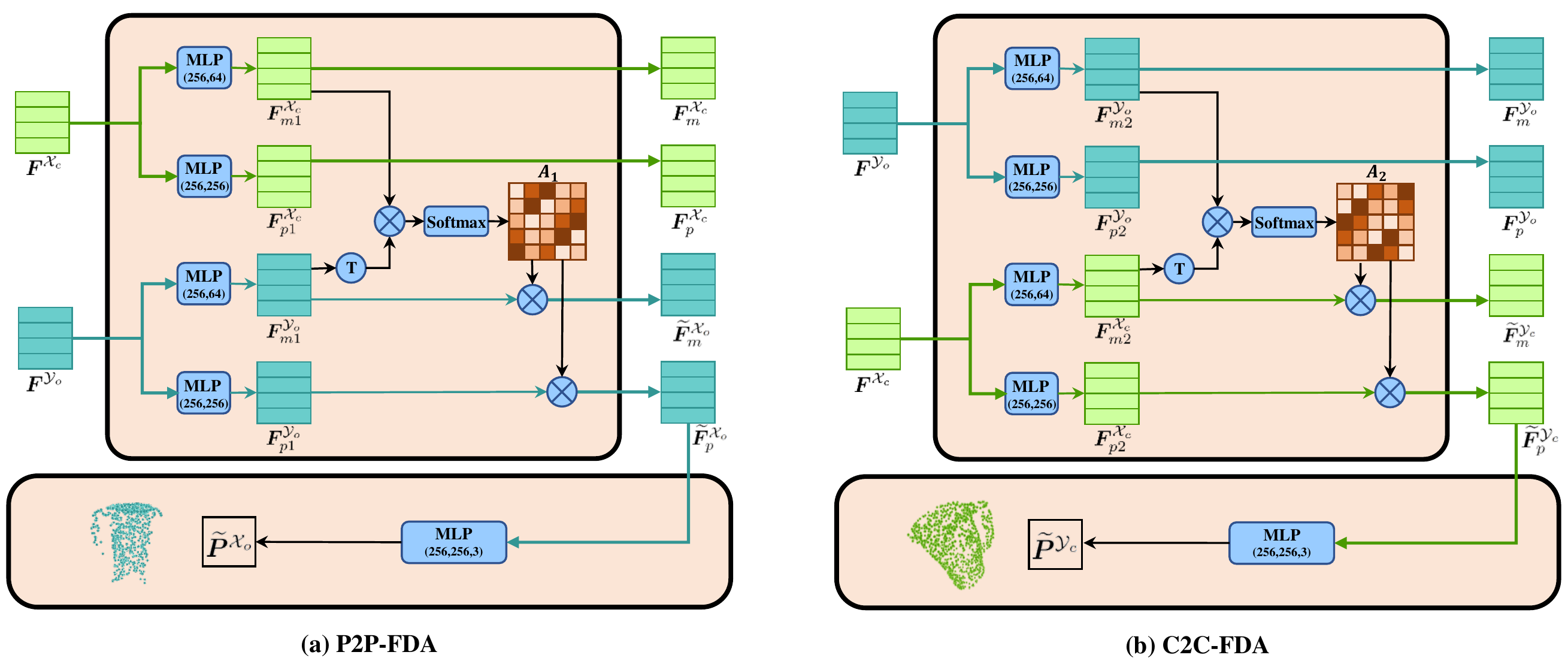}
    \caption{Illustrations of dual Feature Disengagement and Alignment modules. ``T" denotes matrix transposition, and ``$\times$" denotes matrix multiplication. 
    }
    \label{fig:FDA}
\end{figure}

\subsubsection{Feature Disengagement and Alignment Module}

Feature Disengagement and Alignment (FDA) module takes point-wise feature maps of different coordinate systems as inputs, disengages each feature map into pose and match ones, which are then aligned across systems to establish deep correspondence. Fig. \ref{fig:FDA} gives illustrations of both P2P-FDA and C2C-FDA modules, where network specifics are also given.

We take P2P-FDA module (\ref{EqnPartialFDA}) as an example to illustrate the implementation of FDA.
Specifically, as shown in Fig. \ref{fig:FDA}, we firstly disengage  $\bm{F}^{\mathcal{X}_c}$ into a pose feature $\bm{F}^{\mathcal{X}_c}_{p1}$ and a match one $\bm{F}^{\mathcal{X}_c}_{m1}$:
\begin{equation}
\bm{F}^{\mathcal{X}_c}_{p1} = \texttt{MLP}(\bm{F}^{\mathcal{X}_c}), \bm{F}^{\mathcal{X}_c}_{m1} = \texttt{MLP}(\bm{F}^{\mathcal{X}_c}),
\end{equation}
where \texttt{MLP}$(\cdot)$ denotes a subnetwork of Multi-layer Perceptron (MLP).
The same applies to $\bm{F}^{\mathcal{Y}_{o}}$, and we have $\bm{F}^{\mathcal{Y}_{o}}_{p1}$ and $\bm{F}^{\mathcal{Y}_{o}}_{m1}$. The match features $\bm{F}^{\mathcal{X}_c}_{m1}$ and $\bm{F}^{\mathcal{Y}_{o}}_{m1}$ is then used for the learning of an attention map $\bm{A}_1 \in \mathbb{R}^{N_{\mathcal{X}}\times N_{\mathcal{Y}}}$ as follows:
\begin{equation}
  \bm{A}_1 = \texttt{Softmax}(\bm{F}^{\mathcal{X}_c}_{m1} \times \texttt{Transpose}(\bm{F}^{\mathcal{Y}_{o}}_{m1})),
\end{equation}
where \texttt{Transpose}$(\cdot)$ denotes tensor transposition, and \texttt{Softmax}$(\cdot)$ denotes softmax operation along columns. Each element $a_{1,ij}$ in $\bm{A}_1$ indicates the match degree between $i^{th}$ point in $\bm{P}^{\mathcal{X}_c}$ and $j^{th}$ one in $\bm{P}^{\mathcal{Y}_o}$. Then pose and match features of the partial observation $\mathcal{X}_o$ in object system can be interpolated by matrix multiplication of $\bm{A}_1$ and those of $\bm{P}^{\mathcal{Y}_o}$, respectively, to be aligned with features of $\mathcal{X}_c$ in camera coordinate system:
\begin{equation}
  \left\{\begin{array}{l}
    \bm{F}^{\mathcal{X}_{c}}_p =  \bm{F}^{\mathcal{X}_c}_{p1} \\
    \widetilde{\bm{F}}^{\mathcal{X}_{o}}_p =  \bm{A}_1 \times \bm{F}^{\mathcal{Y}_{o}}_{p1}
   \end{array}\right.,
   \left\{\begin{array}{l}
    \bm{F}^{\mathcal{X}_{c}}_m =  \bm{F}^{\mathcal{X}_c}_{m1} \\
    \widetilde{\bm{F}}^{\mathcal{X}_{o}}_m =  \bm{A}_1 \times \bm{F}^{\mathcal{Y}_{o}}_{m1}
   \end{array}\right..
\end{equation}
Through feature alignment, $\widetilde{\bm{P}}^{\mathcal{X}_o}$ is expected to be decoded out from $\widetilde{\bm{F}}^{\mathcal{X}_{o}}_p$:
\begin{equation}
\widetilde{\bm{P}}^{\mathcal{X}_o} = \texttt{MLP}(\widetilde{\bm{F}}^{\mathcal{X}_{o}}_p).
\end{equation}
Supervisions on the reconstruction of $\widetilde{\bm{P}}^{\mathcal{X}_o}$ guide the learning of deep correspondence in P2P-FDA module.

P2P-FDA module (\ref{EqnPartialFDA}) learns deep correspondence of the partial $\mathcal{X}$ in two coordinate systems, while C2C-FDA module (\ref{EqnCompleteFDA}) infers that of the complete $\mathcal{Y}$ via a same network structure, as shown in Fig. \ref{fig:FDA}(b). We adopt dual FDA modules in our design to enable robust correspondence establishment, since they bring complementary functions: P2P-FDA module provides more accurate correspondence than that of C2C-FDA module, due to the difficulty in shape completion from partial observation for the latter module; however, C2C-FDA module plays a vital role under the condition of severe occlusions, which P2P-FDA module can hardly handle with.

\subsection{Confidence-based Pose Estimation}
\label{subsec:PoseEstimation}
After dual feature disengagement and alignment, we construct the pose and match feature pairs as follows:
\begin{equation}
  \bm{F}_p = \begin{bmatrix}
    \bm{F}^{\mathcal{X}_c}_p, & \widetilde{\bm{F}}^{\mathcal{X}_o}_p\\
    \widetilde{\bm{F}}^{\mathcal{Y}_c}_p,& \bm{F}^{\mathcal{Y}_o}_p
   \end{bmatrix},  \bm{F}_m = \begin{bmatrix}
    \bm{F}^{\mathcal{X}_c}_m, & \widetilde{\bm{F}}^{\mathcal{X}_o}_m\\
    \widetilde{\bm{F}}^{\mathcal{Y}_c}_m,& \bm{F}^{\mathcal{Y}_o}_m
   \end{bmatrix}.
\end{equation}
As shown in Fig. \ref{fig:overview}, the paired match feature $\bm{F}_m$ is fed into an MLP for the learning of confidence scores $\bm{s}=\{s_i\}_{i=1}^{N_{\mathcal{X}}+N_{\mathcal{Y}}}$ to reflect the qualities of deep correspondence:
\begin{equation}
  \bm{s} = \texttt{MLP}(\bm{F}_m).
\end{equation}

The paired pose feature $\bm{F}_p$ is also fed into an MLP and weighted by $\bm{s}$ for precisely estimating the 6D pose $(\bm{R}, \bm{t})$:
\begin{eqnarray}
  &\bm{R} = \texttt{MLP}(\bm{f}), \bm{t} = \texttt{MLP}(\bm{f}), \\
  \textit{s.t.} &\bm{f} = \texttt{SUM}(\texttt{SoftMax}(\bm{s}) \cdot \texttt{MLP}(\bm{F}_p)), \notag
\end{eqnarray}
where \texttt{SUM} denotes summation along rows.

Rather than numerical calculation from two paired point sets, we directly regress the 6D object pose from deep pair-wise features with confidence scores, which effectively weakens the negative impact of correspondence of low quality on pose estimation, and thus realizes more precise results.

\subsection{Training of Deep Correspondence Learning Network}
\label{subsec:training_objectives}

For dual FDA modules, we supervise the reconstruction of $\widetilde{\bm{P}}^{\mathcal{X}_o}=\{\widetilde{\bm{p}}_i^{\mathcal{X}_o}\}_{i=1}^{N_{\mathcal{X}}}$ and $\widetilde{\bm{P}}^{\mathcal{Y}_c}=\{\widetilde{\bm{p}}_i^{\mathcal{Y}_c}\}_{i=1}^{N_{\mathcal{Y}}}$ to guide the learning of deep correspondence via the following objectives:
\begin{equation}
\mathcal{L}_{p2p} = \frac{1}{N_{\mathcal{X}}} \sum_{i=1}^{N_{\mathcal{X}}} || \widetilde{\bm{p}}_i^{\mathcal{X}_o} - \bm{R}^{*T}(\bm{p}_i^{\mathcal{X}_c}-\bm{t}^*) ||,
\label{EqnLossX}
\end{equation}
\begin{equation}
\mathcal{L}_{c2c} = \frac{1}{N_{\mathcal{Y}}} \sum_{i=1}^{N_{\mathcal{Y}}} || \widetilde{\bm{p}}_i^{\mathcal{Y}_c} - (\bm{R}^*\bm{p}_i^{\mathcal{Y}_o}+\bm{t}^*) ||,
\label{EqnLossY}
\end{equation}
where $\bm{P}^{\mathcal{X}_c}=\{\bm{p}_i^{\mathcal{X}_c}\}_{i=1}^{N_{\mathcal{X}}}$ and $\bm{P}^{\mathcal{Y}_o}=\{\bm{p}_i^{\mathcal{Y}_o}\}_{i=1}^{N_{\mathcal{Y}}}$ are input point sets, and $\bm{R}^*$ and $\bm{t}^*$ denote ground truth 6D pose.
For the confidence-based pose estimation, we use the following objectives on top of the learning of the predicted object pose $(\bm{R}, \bm{t})$ and confidence scores $\bm{s}=\{s_i\}_{i=1}^{N_{\mathcal{X}}+N_{\mathcal{Y}}}$, respectively:
\begin{equation}
  \mathcal{L}_{pose} = \frac{1}{N_{\mathcal{Y}}} \sum_{i=1}^{N_{\mathcal{Y}}} ||\bm{R}\bm{p}_i^{\mathcal{Y}_o}+\bm{t}- (\bm{R}^*\bm{p}_i^{\mathcal{Y}_o}+\bm{t}^*) ||.
\label{EqnLossPose}
\end{equation}
\begin{eqnarray}
  \mathcal{L}_{conf} = \frac{1}{N_{\mathcal{X}}} \sum_{i=1}^{N_{\mathcal{X}}} \sigma(|| \widetilde{\bm{p}}_i^{\mathcal{X}_o} - \bm{R}^{T}(\bm{p}_i^{\mathcal{X}_c}-\bm{t})  ||,  s_i)  \notag \\
    + \frac{1}{N_{\mathcal{Y}}} \sum_{j=1}^{N_{\mathcal{Y}}} \sigma(|| \widetilde{\bm{p}}_j^{\mathcal{Y}_c} - (\bm{R}\bm{p}_j^{\mathcal{Y}_o}+\bm{t})|| ,  s_{N_{\mathcal{X}}+j}),
\label{EqnLossConf}
\end{eqnarray}
where $\sigma(d,s)=ds-wlog(s)$, and $w$ is a balancing hyperparameter. We note that the objectives (\ref{EqnLossX}), (\ref{EqnLossY}) and (\ref{EqnLossPose}) are designed for asymmetric objects, while for symmetric ones, we modify them by replacing $L_2$ distance with Chamfer distance, as done in \cite{densefusion}.

The overall training objective combines (\ref{EqnLossX}), (\ref{EqnLossY}), (\ref{EqnLossPose}), and (\ref{EqnLossConf}), resulting in the following optimization problem:
\begin{equation}
  \min \mathcal{L} =  \lambda_1 \mathcal{L}_{p2p} + \lambda_2 \mathcal{L}_{c2c} + \lambda_3 \mathcal{L}_{pose} + \lambda_4 \mathcal{L}_{conf},
\label{EqnLossALL}
\end{equation}
where $\lambda_1$, $\lambda_2$, $\lambda_3$ and $\lambda_4$ are penalty parameters. 

\subsection{Confidence-based Pose Refinement}
\label{subsec:confidence_based_pose_refinement}

\begin{figure}[t]
  \centering
  \includegraphics[width=0.99\linewidth]{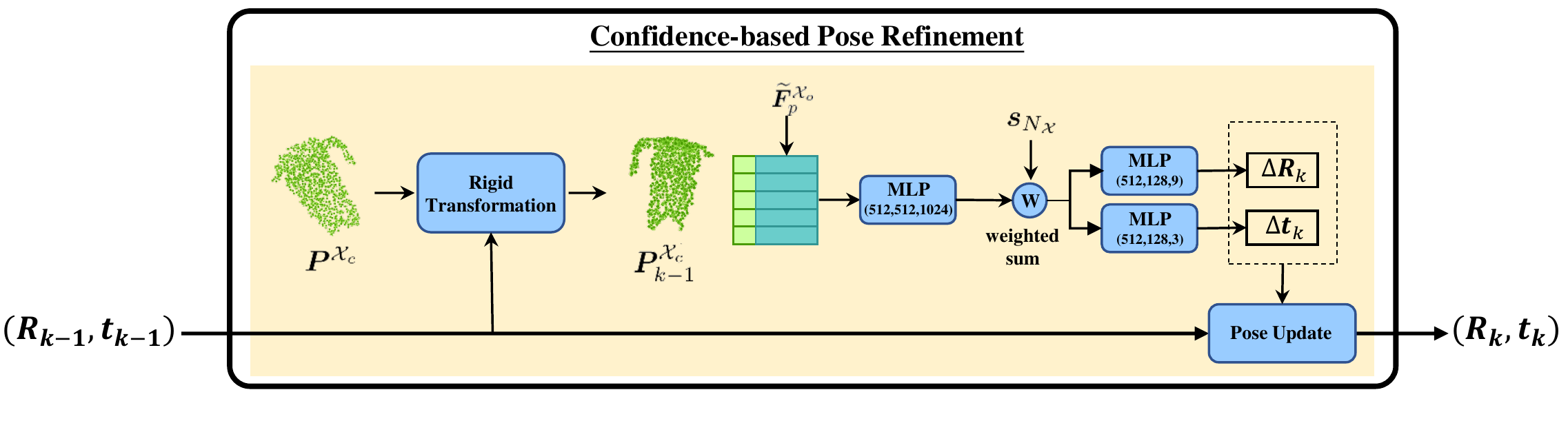}
    \caption{An illustration of the iterative confidence-based pose estimation network. 
    }
    \label{fig:refinement}
\end{figure}

To take full advantages of the learned correspondence, we propose a confidence-based pose refinement network, as shown in Fig. \ref{fig:refinement}, where the input point set $\bm{P}^{\mathcal{X}_c}$ is transformed with predicted pose, and paired with $\widetilde{\bm{F}}^{\mathcal{X}_o}_p$ for residual pose estimation in an iterative manner. Specifically, assuming after $k-1$ iterations of refinement, the current object pose is updated as $(\bm{R}_{k-1}, \bm{t}_{k-1})$, and we use it for transforming $\bm{P}^{\mathcal{X}_c}=\{\bm{p}_i^{\mathcal{X}_c}\}_{i=1}^{N_{\mathcal{X}}}$ to $\bm{P}_{k-1}^{\mathcal{X}_c}=\{\bm{R}_{k-1}^{T}(\bm{p}_i^{\mathcal{X}_c}-\bm{t}_{k-1})\}_{i=1}^{N_{\mathcal{X}}}$; for forming pair-wise pose features with the learned correspondence in dual FDA modules, we reuse $\widetilde{\bm{F}}^{\mathcal{X}_o}_p$ by concatenating it with $\bm{P}_{k-1}^{\mathcal{X}_c}$. Similarly to Sec. \ref{subsec:PoseEstimation}, we feed the pose feature pairs into an MLP, and weight them by reusing the confidence scores  $\bm{s}_{N_\mathcal{X}}$ (denoting the first $N_\mathcal{X}$ elements of $\bm{s}$) for estimating the residual pose $(\Delta\bm{R}_{k}, \Delta\bm{t}_{k})$:
\begin{eqnarray}
  &\Delta\bm{R}_{k} = \texttt{MLP}(\bm{f}_k), \Delta\bm{t}_{k} = \texttt{MLP}(\bm{f}_k), \\
  \textit{s.t.} &\bm{f}_k = \texttt{SUM}(\texttt{SoftMax}(\bm{s}_{N_\mathcal{X}}) \cdot \texttt{MLP}([\bm{P}_{k-1}^{\mathcal{X}_c}, \widetilde{\bm{F}}^{\mathcal{X}_o}_p])). \notag
\end{eqnarray}
Finally, the pose $(\bm{R}_{k}, \bm{t}_{k})$ of the $k^{th}$ iteration can be obtained as follows:
\begin{equation}
  \bm{R}_k = \Delta\bm{R}_k \bm{R}_{k-1}, \bm{t}_k = \bm{R}_{k-1}\Delta\bm{t}_{k} + \bm{t}_{k-1}.
\end{equation}

\section{Experiments}
\label{exp:Exp}

\subsubsection{Datasets} We conduct experiments on three benchmarking datasets, including YCB-Video \cite{YCB}, LineMOD \cite{LineMod}, and Occlusion-LineMOD \cite{learning6d}. 
YCB-Video dataset consists of $92$ RGB-D videos with $21$ different object instances, fully annotated with object poses and masks. Following \cite{densefusion}, we use $80$ videos therein for training along with additional $80,000$ synthetic images, and evaluate DCL-Net on $2,949$ keyframes sampled from the rest $12$ videos. 
LineMOD is also a fully annotated dataset for 6D pose estimation, containing $13$ videos with $13$ low-textured object instances; we follow the prior work \cite{densefusion} to split training and testing sets.
Occlusion-LineMOD is an annotated subset of LineMOD with 8 different object instances, which handpicks RGB-D images of scenes with heavy object occlusions and self-occlusions from LineMOD, making the task of pose estimation more challenging; following \cite{hybridpose}, we use the DCL-Net trained on the original LineMOD to evaluate on Occlusion-LineMOD.
 
\subsubsection{Implementation Details} For both object observations and CAD models, we sample point sets with $1,024$ points as inputs of DCL-Net; that is, $N_{\mathcal{X}}=N_{\mathcal{Y}}=1,024$. For the training objectives, we set the penalty parameters $\lambda_1$, $\lambda_2$, $\lambda_3$, $\lambda_4$ in (\ref{EqnLossALL}) as $5.0$, $1.0$, $1.0$, and $1.0$, respectively; $w$ in (\ref{EqnLossConf}) is set as $0.01$. During inference, we run twice the confidence-based pose refinement for improvement of pose precision.

\subsubsection{Evaluation Metrics} We use the same evaluation metrics as those in \cite{densefusion}. For YCB-Video dataset, the average closest point distance (ADD-S) \cite{posecnn} is employed to measure the pose error; following \cite{densefusion}, we report the Area Under the Curve (AUC) of ADD-S with the maximum threshold at $0.1$m, and the percentage of ADD-S smaller than the minimum tolerance at $2$cm ($<2$cm). For both LineMOD and Occlusion-LineMOD datasets, ADD-S is employed only for symmetric objects, while the Average Distance (ADD) for asymmetric objects; we report the percentage of distance smaller than $10\%$ of object diameter. Besides, we use Chamfer Distance (CD) to measure the reconstruction results.

\subsection{Ablation Studies and Analyses}

We firstly conduct ablation studies to evaluate the efficacy of novel designs proposed in our DCL-Net. These experiments are conducted on YCB-Video dataset \cite{YCB}.

\begin{table}[t]
  \begin{center}
    \caption{Ablation studies of the use of dual FDA modules on YCB-Video dataset \cite{YCB}. Experiments are conducted without confidence-based weighting and pose refinement. }
    \label{tab:AblationFDA}
      \begin{tabular}{M{2cm}|M{2cm}|M{1.3cm}M{1.3cm}M{1.3cm}M{1.3cm}}
        \hline
        \multirow{2}{*}{P2P-FDA} & \multirow{2}{*}{C2C-FDA} &  \multirow{2}{*}{AUC}  &  \multirow{2}{*}{$<2$cm} & \multicolumn{2}{c}{CD ($\times {10}^{-3}$)}\\
        \cline{5-6}
        & & & & $\bm{P}^{\mathcal{X}_o}$ & $\bm{P}^{\mathcal{Y}_c}$ \\
        \hline
        $\times$ & $\times$ & 94.1 & 97.4 & $-$ & $-$\\
        \checkmark & $\times$  & 95.0 & 98.7 & 7.1 & $-$\\
        $\times$ & \checkmark  & 94.5 & 98.8 & $-$ & 8.2\\
        \checkmark & \checkmark  & $\mathbf{95.3}$ & $\mathbf{99.0}$& $\mathbf{7.0}$ & $\mathbf{8.1}$ \\
        \hline
      \end{tabular}
    \end{center}
\end{table}

\subsubsection{Effects of Dual Feature Disengagement and Alignment}  
\begin{figure}[t]
  \centering
  \includegraphics[width=1.0\linewidth]{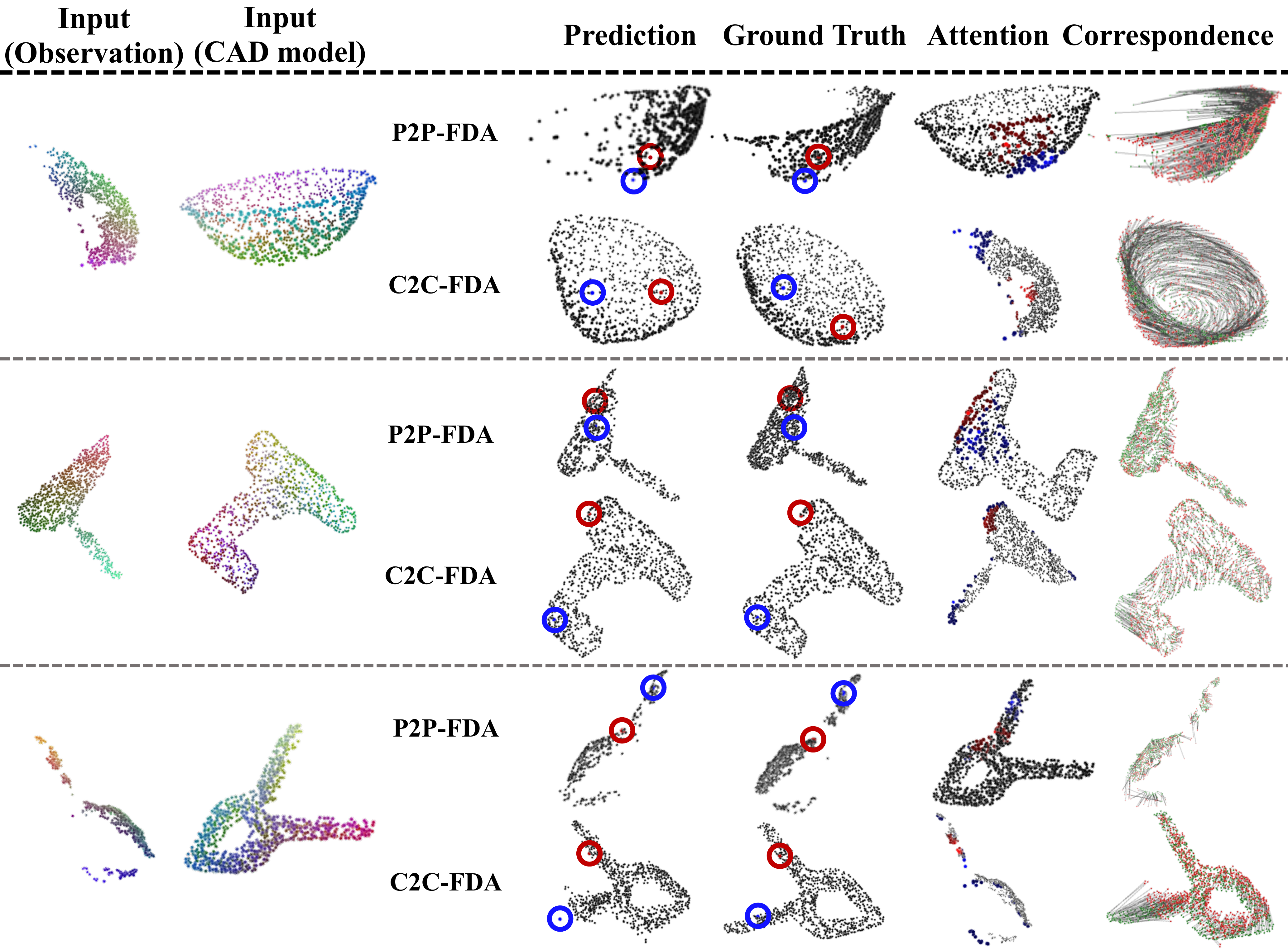}
    \caption{Visualizations of shape predictions, attentions, and correspondence of both P2P-FDA and C2C-FDA modules on YCB-Video dataset \cite{YCB}. Best view in electronic version.}
    \label{fig:resutls_fda}
\end{figure}

We conduct four experiments to evaluate the efficacy of the use of dual FDA modules: i) without any FDA modules (baseline), ii) only with P2P-FDA, iii) only with C2C-FDA, and iv) with dual modules. For simplicity, these experiments are conducted without confidence-based weighting as well as pose refinement. The quantitative results on ADD-S AUC and ADD-S$<2$cm are shown in Table \ref{tab:AblationFDA}, where the reconstruction results of asymmetric objects are also reported. From the table, methods with (one or dual) FDA modules indeed outperforms the baseline, which demonstrates the importance of deep correspondence learning on pose estimation. Single P2P-FDA module achieves more accurate results than single C2C-FDA module by making better reconstructions ($7.1\times 10^{-3}$ versus $8.2 \times 10^{-3}$ on CD) and deep correspondence as well, and the mixed use of them boosts the performance, indicating their complementary advantages. For the last framework, we visualize the reconstruction results along with the learned correspondence of both P2P-FDA and C2C-FDA modules in Fig. \ref{fig:resutls_fda}; shape completion can be achieved for C2C-FDA module, even with severe occlusions, to build valid deep correspondence of high quality, and thus make DCL-Net more robust and reliable. 

\begin{table}[t]
  \begin{center}
    \caption{Quantitative results obtained by least-squares optimization \cite{arun1987least} and our proposed direct regression on YCB-Video dataset \cite{YCB}. Experiments are conducted without pose refinement. }
    \label{tab:AblationConf}
      \begin{tabular}{M{2cm}|M{5cm}|M{1.5cm}M{1.5cm}}
      \hline
        &  & AUC & $<2$cm  \\
      \hline
      \multirow{2}{*}{w/o Conf.} & Least-squares Optimization \cite{arun1987least} & 94.7 & 98.2 \\
      & Direct Pose Regression & 95.3 & 99.0 \\
      \hline
      \multirow{2}{*}{with Conf.} & Least-squares Optimization \cite{arun1987least} & 95.4 & 98.3\\
      & Direct Pose Regression & $\mathbf{95.8}$ & $\mathbf{99.0}$ \\
      \hline 
      \end{tabular}
    \end{center}
\end{table}

We also explore the attention maps of dual FDA modules in Fig. \ref{fig:resutls_fda}. Take C2C-FDA module as an example, the predicted points are learned from the features of the input observed ones via attention maps, \textit{i.e.}, each predicted point corresponds to the observed ones with different attention weights, and we thus colorize those corresponding points with large weights in Fig. \ref{fig:resutls_fda}; as shown in the figure, for the predicted points (red) locate at the observed parts, most of the input points with larger weights (red) could locate at the corresponding local regions, showing the qualities of attention maps, while for those at the occluded parts (blue), the corresponding points (blue) may locate scatteredly, but thanks to the correspondence learning in feature space, these points could still be completed in the C2C-FDA reconstruction results.

\subsubsection{Effects of Confidence-based Pose Estimation} Through learning deep correspondence in feature space, DCL-Net achieves direct regression of object poses, while the predictions of dual FDA modules can also establish point correspondence \textit{w.r.t} inputs to solve poses via least-squares optimization \cite{arun1987least}. We compare the quantitative results obtained by these two approaches (without pose refinement) in Table \ref{tab:AblationConf}, where results of direct regression from deep feature correspondence outperforms those from point correspondence consistently with or without confidence scores,  showing that pose estimation from feature space is less sensitive to the correspondence of low qualities, thanks to the direct objectives for the end task. Besides, we also observe that the learning of confidence scores not only measures the qualities of correspondence and decreases the influence of bad correspondence, but also helps improve the qualities themselves effectively.

\begin{table}[t]
  \caption{Quantitative results of DCL-Net with or without pose refinement on YCB-Video dataset \cite{YCB}. }
  \label{tab:AblationRefine}
  \begin{center}
      \begin{tabular}{M{4cm}|M{1.5cm}M{1.5cm}}
        \hline
          & AUC & $<2$cm \\
        \hline
        w/o Pose Refinement  & 95.8 & 99.0\\
        with Pose Refinement  & $\mathbf{96.6}$ & $\mathbf{99.0}$\\
        \hline
      \end{tabular}
    \end{center}
\end{table}

\begin{figure}[t]
  \centering
  \includegraphics[width=1.0\linewidth]{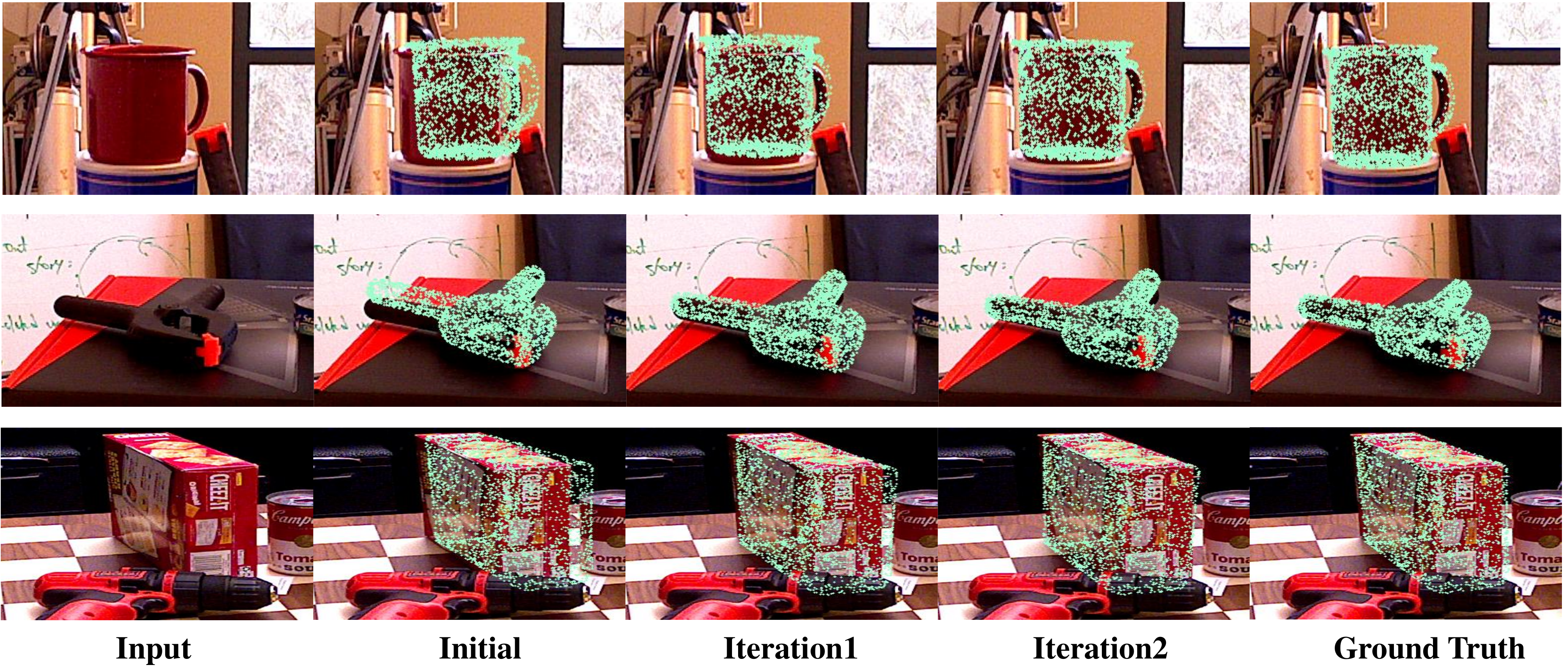}
    \caption{Qualitative results of DCL-Net with or without pose refinement on YCB-Video dataset \cite{YCB}. The sampled points of CAD models are transformed by the predicted poses and projected to 2D images.}
    \label{fig:results_refinement}
\end{figure}

\subsubsection{Effects of Confidence-based Pose Refinement} Table \ref{exp:sota_ycb} demonstrates the efficiency of our confidence-based pose refinement for boosting the performance, \eg, improvement by $0.8\%$ on the metric of ADD-S AUC, which is also verified by the qualitative results shown in Fig. \ref{fig:results_refinement}.

\subsection{Comparisons with Existing Methods}

\begin{table}[t]
  \centering
  \caption{Quantitative results of different methods on YCB-Video dataset \cite{YCB}. The evaluation metrics are ADD-S AUC and ADD-S$<2$cm. Objects with bold name are symmetric. }
  \label{exp:sota_ycb}
  \resizebox{0.99\textwidth}{!}{
  \begin{tabular}{l|cc|cc|c|cc|cc|cc|cc}
  \toprule
                           & \multicolumn{2}{c|}{PoseCNN+ICP\cite{posecnn}} & \multicolumn{2}{c|}{DenseFusion\cite{densefusion}} & G2L\cite{G2L}   & \multicolumn{2}{c|}{PVN3D\cite{pvn3d}} & \multicolumn{2}{c|}{PR-GCN\cite{PRGCN}}  & \multicolumn{2}{c|}{FFB6D}          & \multicolumn{2}{c}{DCL-Net}    \\
                           & AUC       & $\textless$2cm                     & AUC        & $\textless$2cm                        & AUC             & AUC              & $\textless$2cm      & AUC              & $\textless$2cm        & AUC             & $\textless$2cm    & AUC     & $\textless$2cm   \\
  \midrule 
  002\_master\_chef\_can   & 95.8      & \textbf{100.0}                     & 96.4       & \textbf{100.0}                        & 94.0            & 96.0             & \textbf{100.0}      & \textbf{97.1}    & \textbf{100.0}        & 96.3            & \textbf{100.0}       & 96.1            & \textbf{100.0} \\
  003\_cracker\_box        & 92.7      & 91.6                               & 95.5       & 99.5                                  & 88.7            & 96.1             & \textbf{100.0}      & \textbf{97.6}    & \textbf{100.0}        & 96.3            & \textbf{100.0}                & 96.4            & 99.4           \\
  004\_sugar\_box          & 98.2      & \textbf{100.0}                     & 97.5       & \textbf{100.0}                        & 96.0            & 97.4             & \textbf{100.0}      & \textbf{98.3}    & \textbf{100.0}        & 97.6            & \textbf{100.0}       & 98.1            & \textbf{100.0} \\
  005\_tomato\_soup\_can   & 94.5      & 96.9                               & 94.96      & 96.9                                  & 86.4            & \textbf{96.2}    & \textbf{98.1}       & 95.3             & 97.6                  & 95.6            & 98.2                           & 95.8            & 97.7           \\
  006\_mustard\_bottle     & 98.6      & \textbf{100.0}                     & 97.2       & \textbf{100.0}                        & 95.9            & 97.5             & \textbf{100.0}      & 97.9             & \textbf{100.0}        & 97.8            & \textbf{100.0}       & \textbf{98.7}   & \textbf{100.0} \\
  007\_tuna\_fish\_can     & 97.1      & \textbf{100.0}                     & 96.6       & \textbf{100.0}                        & 84.1            & 96.0             & \textbf{100.0 }     & 97.6             & \textbf{100.0}        & 96.8            & \textbf{100.0}       & 97.4            & \textbf{100.0} \\
  008\_pudding\_box        & 97.9      & \textbf{100.0}                     & 96.5       & \textbf{100.0}                        & 93.5            & 97.1             & \textbf{100.0}      & \textbf{98.4}    & \textbf{100.0}        & 97.1            & \textbf{100.0}       & 98.2            & \textbf{100.0} \\
  009\_gelatin\_box        & 98.8      & \textbf{100.0}                     & 98.1       & \textbf{100.0}                        & 96.8            & 97.7             & \textbf{100.0}      & 96.2             & 94.4                  & 98.1            & \textbf{100.0}    & \textbf{98.9}   & \textbf{100.0} \\
  010\_potted\_meat\_can   & 92.7      & 93.6                               & 91.3       & 93.1                                  & 86.2            & 93.3             & 94.6                & \textbf{96.6}    & \textbf{99.1}         & 94.7            & 94.3                   & 93.1            & 94.7           \\
  011\_banana              & 97.1      & 99.7                               & 96.6       & \textbf{100.0}                        & 96.3            & 96.6             & \textbf{100.0}      & \textbf{98.5}    & \textbf{100.0}        & 97.2            & \textbf{100.0}     & 98.1            & \textbf{100.0} \\
  019\_pitcher\_base       & 97.8      & \textbf{100.0}                     & 97.1       & \textbf{100.0}                        & 91.8            & 97.4             & \textbf{100.0}      & 98.1             & \textbf{100.0}        & 97.6            & \textbf{100.0}           & 98.0            & 99.8            \\
  021\_bleach\_cleanser    & 96.9      & 99.4                               & 95.8       & \textbf{100.0}                        & 92.0            & 96.0             & \textbf{100.0}      & \textbf{97.9}    & \textbf{100.0}        & 96.8            & \textbf{100.0}       & 97.0            & \textbf{100.0} \\
  \textbf{024\_bowl}       & 81.0      & 54.9                               & 88.2       & 98.8                                  & 86.7            & 90.2             & 80.5                & 90.3             & 96.6                  & 96.3            & \textbf{100.0}        & \textbf{97.3}   & \textbf{100.0} \\
  025\_mug                 & 95.0      & 99.8                               & 97.1       & \textbf{100.0}                        & 95.4            & 97.6             & \textbf{100.0}      & \textbf{98.1}    & \textbf{100.0}        & 97.3            & \textbf{100.0}             & 97.8            & \textbf{100.0} \\
  035\_power\_drill        & \textbf{98.2}      & 99.6                      & 96.0       & 98.7                                  & 95.2            & 96.7             & \textbf{100.0}      & 98.1             & \textbf{100.0}        & 97.2            & \textbf{100.0}        & 98.0            & \textbf{100.0} \\
  \textbf{036\_wood\_block}& 87.6      & 80.2                               & 89.7       & 94.6                                  & 86.2            & 90.4             & 93.8                & \textbf{96.0}    & \textbf{100.0}        & 92.6            & 92.1                 & 93.9            & 97.5           \\
  037\_scissors            & 91.7      & 95.6                               & 95.2       & \textbf{100.0 }                       & 83.8            & 96.7             & \textbf{100.0}      & \textbf{96.7}    & \textbf{100.0  }      & 97.7            & \textbf{100.0}       & 87.6            & 98.3           \\
  040\_large\_marker       & 97.2      & 99.7                               & 97.5       & \textbf{100.0   }                     & 96.8            & 96.7             & 99.8                & 97.9             & \textbf{100.0}        & 96.6            & \textbf{100.0}      & 97.8            & 99.8           \\
  \textbf{051\_large\_clamp}        & 75.2      & 74.9                      & 72.9       & 79.2                                  & 94.4            & 93.6             & 93.6                & 87.5             & 93.3                  & 96.8            & \textbf{100.0}         & 95.7            & 98.6           \\
  \textbf{052\_extra\_large\_clamp} & 64.4      & 48.8                      & 69.8       & 76.3                                  & \textbf{92.3}   & 88.4             & 83.6                & 79.7             & 84.6                  & \textbf{96.0}   & \textbf{98.6}        & 88.8            & 87.2           \\
  \textbf{061\_foam\_brick}         & 97.2      & \textbf{100.0}            & 92.5       & \textbf{100.0}                        & 94.7            & 96.8             & \textbf{100.0}      & \textbf{97.8}    & \textbf{100.0}        & 97.3            & \textbf{100.0}        & 97.5            & \textbf{100.0} \\
  \midrule         
  MEAN                     & 93.0      & 93.2                               & 93.1       & 96.8                                  & 92.4            & 95.5             & 97.6                & 95.8             & 98.5                  & \textbf{96.6}   & \textbf{99.2}         & \textbf{96.6}   & 99.0           \\       
  \bottomrule

  \end{tabular}}
\end{table}

\begin{figure}[!h]
  \centering
  \includegraphics[width=1.0\linewidth]{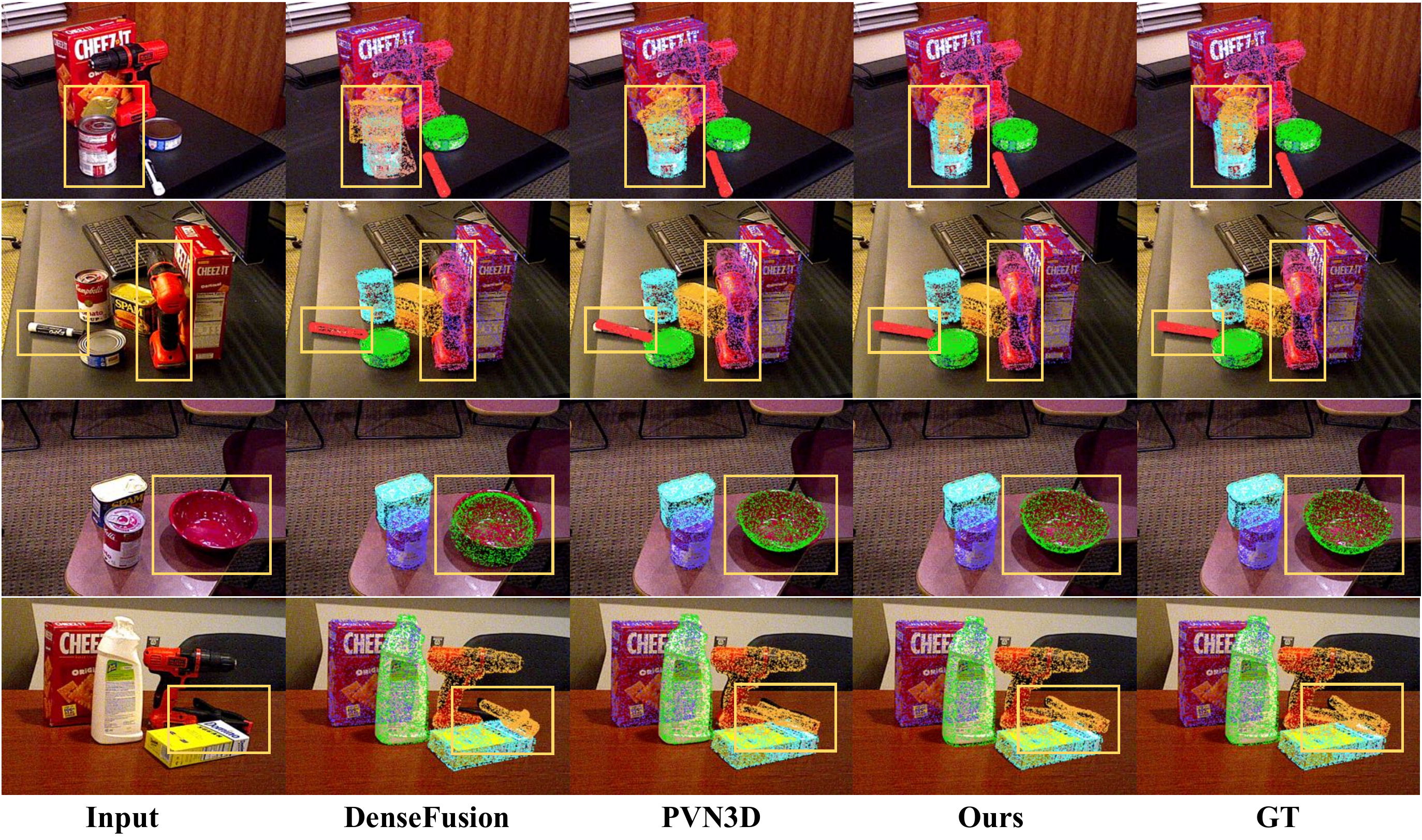}
    \caption{Qualitative results of different methods on YCB-Video dataset \cite{YCB}. The sampled points of CAD models are transformed by the predicted poses and projected to 2D images.}
    \label{fig:exp_ycb}
\end{figure}

We compare our proposed DCL-Net with the existing methods for 6D object pose estimation from RGB-D data, including those based on direct regression (\eg, DenseFusion \cite{densefusion} and PR-GCN \cite{PRGCN}), and those based on dense correspondence learning (\eg, PVN3D \cite{pvn3d} and FFB6D\cite{FFB6D}). 

Quantitative results on the three benchmarking datasets, including YCB-Video \cite{YCB}, LineMOD \cite{LineMod}, and Occlusion-LineMOD \cite{learning6d}, are reported in Table \ref{exp:sota_ycb}, Table \ref{exp:sota_linemod}, and Table \ref{exp:sota_LMO}, respectively, all of which show the superiority of our DCL-Net consistently in the regime of pose precision;  qualitative results on YCB-Video dataset \cite{YCB} are also provided in Fig. \ref{fig:exp_ycb} to verify the advantages of our DCL-Net. Remarkably, on the more challenging Occlusion-LineMOD dataset, the improvements of our DCL-Net over the state-of-the-art methods of PR-GCN \cite{PRGCN} and FFB6D \cite{FFB6D} reach $5.6\%$ and $4.4\%$ on the metric of ADD(S), respectively, indicating the advantages of our DCL-Net on handling with object occlusions or self-occlusions.

\begin{table}[t]
  \centering
  \caption{Quantitative results of different methods on ADD(S) on LineMOD dataset \cite{LineMod}. Objects with bold name are symmetric.}
  \label{exp:sota_linemod}
  \resizebox{1.0\textwidth}{!}{
  \begin{tabular}{l|ccccccc|c}
  \toprule
          & \begin{tabular}[c]{@{}c@{}}Implicit\\ +ICP\cite{implicit}\end{tabular} & \begin{tabular}[c]{@{}c@{}}SSD6D\\ +ICP\cite{ssd6d}\end{tabular} & \begin{tabular}[c]{@{}c@{}}PointFusion\\\cite{pointfusion}\end{tabular} & \begin{tabular}[c]{@{}c@{}}DenseFusion\\\cite{densefusion}\end{tabular} & \begin{tabular}[c]{@{}c@{}}DenseFusion\\ (Iterative)\cite{densefusion}\end{tabular} & \begin{tabular}[c]{@{}c@{}}G2L\\\cite{G2L}\end{tabular} & \begin{tabular}[c]{@{}c@{}}PR-GCN\\ \cite{PRGCN}\end{tabular}  & DCL-Net \\
  \midrule
  ape     & 20.6  & 65   & 70.4 & 79.5 & 92.3  & 96.8     & \textbf{97.6}    & 97.4\\
  bench   & 64.3  & 80   & 80.7 & 84.2 & 93.2  & 96.1     & 99.2    & \textbf{99.4}\\
  camera  & 63.2  & 78   & 60.8 & 76.5 & 94.4  & 98.2     & 99.4    & \textbf{99.8}\\
  can     & 76.1  & 86   & 61.1  & 86.6 & 93.1  & 98.0     & 98.4   & \textbf{99.9}\\
  cat     & 72.0  & 70   & 79.1  & 88.8 & 96.5  & 99.2     & 98.7   & \textbf{100.0}\\
  driller & 41.6  & 73 & 47.3   & 77.7  & 87.0  & 99.8     & 98.8   & \textbf{99.9}\\
  duck    & 32.4  & 66   & 63.0  & 76.3  & 92.3 & 97.7     & \textbf{98.9}   & 98.4\\
  \textbf{egg}     & 98.6 & \textbf{100}  & 99.9  & 99.9   & 99.8    & \textbf{100.0}    & 99.9   & \textbf{100.0}\\
  \textbf{glue}    & 96.4   & \textbf{100}  & 99.3 & 99.4  & \textbf{100.0}   & \textbf{100.0}    & \textbf{100.0}        & 99.9\\
  hole    & 49.9   & 49   & 71.8  & 79.0  & 92.1   & 99.0     & 99.4        & \textbf{100.0}\\
  iron    & 63.1  & 78    & 83.2  & 92.1  & 97.0   & 99.3     & 98.5        & \textbf{100.0}\\
  lamp    & 91.7  & 73  & 62.3  & 92.3  & 95.3  & \textbf{99.5}     & 99.2        & \textbf{99.5}\\
  phone   & 71.0  & 79  & 78.8  & 88.0  & 92.8  & 98.9     & 98.4        & \textbf{99.7}\\
  \midrule
  MEAN    & 64.7  & 79  & 73.7  & 86.2 & 94.3   & 98.7     & 98.9       & \textbf{99.5}\\
  \bottomrule
  \end{tabular} }
  \end{table}

\begin{table}[t]
    \centering
    \caption{Quantitative results of different methods on ADD(S) on Occlusion-LineMOD dataset \cite{learning6d}. Objects with bold name are symmetric.}
  \label{exp:sota_LMO}
  \resizebox{1.0\textwidth}{!}{
  \begin{tabular}{l|ccccccccc|c}
  \toprule
                    &  \makecell[c]{PoseCNN \\ \cite{posecnn}} & \makecell[c]{Deep- \\ Heat\cite{deepheat}} & \makecell[c]{SS \\ \cite{ss}}  & \makecell[c]{Pix2pose \\ \cite{pix2pose}} & \makecell[c]{PVNet\\\cite{pvnet}} & \makecell[c]{Hybrid-\\Pose\cite{hybridpose}} & \makecell[c]{PVN3D\\\cite{pvn3d}} & \makecell[c]{PR-GCN\\\cite{PRGCN}}  & \makecell[c]{FFB6D\\\cite{FFB6D}}  & DCL-Net          \\
  \midrule         
  ape              &  9.6                                      & 12.1                                       & 17.6                           & 22.0                                      & 15.8                              & 20.9                                         & 33.9                              & 40.2                                & 47.2                                & \textbf{56.7} \\
  can              &  45.2                                     & 39.9                                       & 53.9                           & 44.7                                      & 63.3                              & 75.3                                         & \textbf{88.6}                     & 76.2                                & 85.2                                & 80.2          \\
  cat              &  0.9                                      & 8.2                                        & 3.3                            & 22.7                                      & 16.7                              & 24.9                                         & 39.1                              & \textbf{57.0}                       & 45.7                                & 48.1          \\
  driller          &  41.4                                     & 45.2                                       & 62.4                           & 44.7                                      & 65.7                              & 70.2                                         & 78.4                              & \textbf{82.3}                       & 81.4                                & 81.4          \\
  duck             &  19.6                                     & 17.2                                       & 19.2                           & 15.0                                      & 25.2                              & 27.9                                         & 41.9                              & 30.0                                & \textbf{53.9}                       & 44.6          \\
  \textbf{egg}     &  22.0                                     & 22.1                                       & 25.9                           & 25.2                                      & 50.2                              & 52.4                                         & 80.9                              & 68.2                                & 70.2                                & \textbf{83.6} \\
  \textbf{glue}    &  38.5                                     & 35.8                                       & 39.6                           & 32.4                                      & 49.6                              & 53.8                                         & 68.1                              & 67.0                                & 60.1                                & \textbf{79.1} \\
  hole             &  22.1                                     & 36.0                                       & 21.3                           & 49.5                                      & 39.7                              & 54.2                                         & 74.7                              & \textbf{97.2}                       & 85.9                                & 91.3          \\
  \midrule                                                                                                
  MEAN             &  24.9                                     & 27.0                                       & 27.0                           & 32.0                                      & 40.8                              & 47.5                                         & 63.2                              & 65.0                                & 66.2                                & \textbf{70.6} \\
  \bottomrule
  \end{tabular}
  }
\end{table}

~\\

\noindent\textbf{Acknowledgements.} This work is supported in part by Guangdong R$\&$D key project of China (No.: 2019B010155001), and the Program for Guangdong Introducing Innovative and Enterpreneurial Teams (No.: 2017ZT07X183).  We also thank Yi Li and Xun Xu for their valuable comments.

\newpage

\bibliographystyle{splncs04}
\bibliography{egbib}

\begin{thebibliography}{10}
\providecommand{\url}[1]{\texttt{#1}}
\providecommand{\urlprefix}{URL }
\providecommand{\doi}[1]{https://doi.org/#1}

\bibitem{arun1987least}
Arun, K.S., Huang, T.S., Blostein, S.D.: Least-squares fitting of two 3-d point
  sets. IEEE Transactions on pattern analysis and machine intelligence (5),
  698--700 (1987)

\bibitem{ICP}
Besl, P.J., McKay, N.D.: Method for registration of 3-d shapes. In: Sensor
  fusion IV: control paradigms and data structures. vol.~1611, pp. 586--606.
  International Society for Optics and Photonics (1992)

\bibitem{learning6d}
Brachmann, E., Krull, A., Michel, F., Gumhold, S., Shotton, J., Rother, C.:
  Learning 6d object pose estimation using 3d object coordinates. In: European
  conference on computer vision. pp. 536--551. Springer (2014)

\bibitem{YCB}
Calli, B., Singh, A., Walsman, A., Srinivasa, S., Abbeel, P., Dollar, A.M.: The
  ycb object and model set: Towards common benchmarks for manipulation
  research. In: 2015 international conference on advanced robotics (ICAR). pp.
  510--517. IEEE (2015)

\bibitem{G2L}
Chen, W., Jia, X., Chang, H.J., Duan, J., Leonardis, A.: G2l-net: Global to
  local network for real-time 6d pose estimation with embedding vector
  features. In: Proceedings of the IEEE/CVF conference on computer vision and
  pattern recognition. pp. 4233--4242 (2020)

\bibitem{FSNet}
Chen, W., Jia, X., Chang, H.J., Duan, J., Shen, L., Leonardis, A.: Fs-net: Fast
  shape-based network for category-level 6d object pose estimation with
  decoupled rotation mechanism. In: Proceedings of the IEEE/CVF Conference on
  Computer Vision and Pattern Recognition. pp. 1581--1590 (2021)

\bibitem{moped}
Collet, A., Martinez, M., Srinivasa, S.S.: The moped framework: Object
  recognition and pose estimation for manipulation. The international journal
  of robotics research  \textbf{30}(10),  1284--1306 (2011)

\bibitem{deng2022vista}
Deng, S., Liang, Z., Sun, L., Jia, K.: Vista: Boosting 3d object detection via
  dual cross-view spatial attention. In: Proceedings of the IEEE/CVF Conference
  on Computer Vision and Pattern Recognition. pp. 8448--8457 (2022)

\bibitem{kitti}
Geiger, A., Lenz, P., Urtasun, R.: Are we ready for autonomous driving? the
  kitti vision benchmark suite. In: 2012 IEEE conference on computer vision and
  pattern recognition. pp. 3354--3361. IEEE (2012)

\bibitem{SparseConv}
Graham, B., Engelcke, M., Van Der~Maaten, L.: 3d semantic segmentation with
  submanifold sparse convolutional networks. In: Proceedings of the IEEE
  conference on computer vision and pattern recognition. pp. 9224--9232 (2018)

\bibitem{discriminative}
Gu, C., Ren, X.: Discriminative mixture-of-templates for viewpoint
  classification. In: European Conference on Computer Vision. pp. 408--421.
  Springer (2010)

\bibitem{SA-SSD}
He, C., Zeng, H., Huang, J., Hua, X.S., Zhang, L.: Structure aware single-stage
  3d object detection from point cloud. In: Proceedings of the IEEE/CVF
  Conference on Computer Vision and Pattern Recognition. pp. 11873--11882
  (2020)

\bibitem{FFB6D}
He, Y., Huang, H., Fan, H., Chen, Q., Sun, J.: Ffb6d: A full flow bidirectional
  fusion network for 6d pose estimation. In: Proceedings of the IEEE/CVF
  Conference on Computer Vision and Pattern Recognition. pp. 3003--3013 (2021)

\bibitem{pvn3d}
He, Y., Sun, W., Huang, H., Liu, J., Fan, H., Sun, J.: Pvn3d: A deep point-wise
  3d keypoints voting network for 6dof pose estimation. In: Proceedings of the
  IEEE/CVF conference on computer vision and pattern recognition. pp.
  11632--11641 (2020)

\bibitem{gradient}
Hinterstoisser, S., Cagniart, C., Ilic, S., Sturm, P., Navab, N., Fua, P.,
  Lepetit, V.: Gradient response maps for real-time detection of textureless
  objects. IEEE transactions on pattern analysis and machine intelligence
  \textbf{34}(5),  876--888 (2011)

\bibitem{LineMod}
Hinterstoisser, S., Holzer, S., Cagniart, C., Ilic, S., Konolige, K., Navab,
  N., Lepetit, V.: Multimodal templates for real-time detection of texture-less
  objects in heavily cluttered scenes. In: 2011 international conference on
  computer vision. pp. 858--865. IEEE (2011)

\bibitem{ss}
Hu, Y., Fua, P., Wang, W., Salzmann, M.: Single-stage 6d object pose
  estimation. In: Proceedings of the IEEE/CVF conference on computer vision and
  pattern recognition. pp. 2930--2939 (2020)

\bibitem{hausdorff}
Huttenlocher, D.P., Klanderman, G.A., Rucklidge, W.J.: Comparing images using
  the hausdorff distance. IEEE Transactions on pattern analysis and machine
  intelligence  \textbf{15}(9),  850--863 (1993)

\bibitem{ssd6d}
Kehl, W., Manhardt, F., Tombari, F., Ilic, S., Navab, N.: Ssd-6d: Making
  rgb-based 3d detection and 6d pose estimation great again. In: Proceedings of
  the IEEE international conference on computer vision. pp. 1521--1529 (2017)

\bibitem{localrgbd}
Kehl, W., Milletari, F., Tombari, F., Ilic, S., Navab, N.: Deep learning of
  local rgb-d patches for 3d object detection and 6d pose estimation. In:
  European conference on computer vision. pp. 205--220. Springer (2016)

\bibitem{levinson2011towards}
Levinson, J., Askeland, J., Becker, J., Dolson, J., Held, D., Kammel, S.,
  Kolter, J.Z., Langer, D., Pink, O., Pratt, V., et~al.: Towards fully
  autonomous driving: Systems and algorithms. In: 2011 IEEE intelligent
  vehicles symposium (IV). pp. 163--168. IEEE (2011)

\bibitem{MCN}
Li, C., Bai, J., Hager, G.D.: A unified framework for multi-view multi-class
  object pose estimation. In: Proceedings of the european conference on
  computer vision (eccv). pp. 254--269 (2018)

\bibitem{independent}
Liebelt, J., Schmid, C., Schertler, K.: independent object class detection
  using 3d feature maps. In: 2008 IEEE Conference on Computer Vision and
  Pattern Recognition. pp.~1--8. IEEE (2008)

\bibitem{SS-Conv}
Lin, J., Li, H., Chen, K., Lu, J., Jia, K.: Sparse steerable convolutions: An
  efficient learning of se (3)-equivariant features for estimation and tracking
  of object poses in 3d space. Advances in Neural Information Processing
  Systems  \textbf{34} (2021)

\bibitem{lin2022category}
Lin, J., Wei, Z., Ding, C., Jia, K.: Category-level 6d object pose and size
  estimation using self-supervised deep prior deformation networks. arXiv
  preprint arXiv:2207.05444  (2022)

\bibitem{dualPoseNet}
Lin, J., Wei, Z., Li, Z., Xu, S., Jia, K., Li, Y.: Dualposenet: Category-level
  6d object pose and size estimation using dual pose network with refined
  learning of pose consistency. In: Proceedings of the IEEE/CVF International
  Conference on Computer Vision. pp. 3560--3569 (2021)

\bibitem{handsAR}
Marchand, E., Uchiyama, H., Spindler, F.: Pose estimation for augmented
  reality: a hands-on survey. IEEE transactions on visualization and computer
  graphics  \textbf{22}(12),  2633--2651 (2015)

\bibitem{hourglass}
Newell, A., Yang, K., Deng, J.: Stacked hourglass networks for human pose
  estimation. In: European conference on computer vision. pp. 483--499.
  Springer (2016)

\bibitem{deepheat}
Oberweger, M., Rad, M., Lepetit, V.: Making deep heatmaps robust to partial
  occlusions for 3d object pose estimation. In: Proceedings of the European
  Conference on Computer Vision (ECCV). pp. 119--134 (2018)

\bibitem{pix2pose}
Park, K., Patten, T., Vincze, M.: Pix2pose: Pixel-wise coordinate regression of
  objects for 6d pose estimation. In: Proceedings of the IEEE/CVF International
  Conference on Computer Vision. pp. 7668--7677 (2019)

\bibitem{pvnet}
Peng, S., Liu, Y., Huang, Q., Zhou, X., Bao, H.: Pvnet: Pixel-wise voting
  network for 6dof pose estimation. In: Proceedings of the IEEE/CVF Conference
  on Computer Vision and Pattern Recognition. pp. 4561--4570 (2019)

\bibitem{rios2013discriminatively}
Rios-Cabrera, R., Tuytelaars, T.: Discriminatively trained templates for 3d
  object detection: A real time scalable approach. In: Proceedings of the IEEE
  international conference on computer vision. pp. 2048--2055 (2013)

\bibitem{localaffine}
Rothganger, F., Lazebnik, S., Schmid, C., Ponce, J.: 3d object modeling and
  recognition using local affine-invariant image descriptors and multi-view
  spatial constraints. International journal of computer vision
  \textbf{66}(3),  231--259 (2006)

\bibitem{orb}
Rublee, E., Rabaud, V., Konolige, K., Bradski, G.: Orb: An efficient
  alternative to sift or surf. In: 2011 International conference on computer
  vision. pp. 2564--2571. Ieee (2011)

\bibitem{hybridpose}
Song, C., Song, J., Huang, Q.: Hybridpose: 6d object pose estimation under
  hybrid representations. In: Proceedings of the IEEE/CVF conference on
  computer vision and pattern recognition. pp. 431--440 (2020)

\bibitem{implicit}
Sundermeyer, M., Marton, Z.C., Durner, M., Brucker, M., Triebel, R.: Implicit
  3d orientation learning for 6d object detection from rgb images. In:
  Proceedings of the European Conference on Computer Vision (ECCV). pp.
  699--715 (2018)

\bibitem{tejani2014latent}
Tejani, A., Tang, D., Kouskouridas, R., Kim, T.K.: Latent-class hough forests
  for 3d object detection and pose estimation. In: European Conference on
  Computer Vision. pp. 462--477. Springer (2014)

\bibitem{SPD}
Tian, M., Ang, M.H., Lee, G.H.: Shape prior deformation for categorical 6d
  object pose and size estimation. In: European Conference on Computer Vision.
  pp. 530--546. Springer (2020)

\bibitem{densefusion}
Wang, C., Xu, D., Zhu, Y., Mart{\'\i}n-Mart{\'\i}n, R., Lu, C., Fei-Fei, L.,
  Savarese, S.: Densefusion: 6d object pose estimation by iterative dense
  fusion. In: Proceedings of the IEEE/CVF conference on computer vision and
  pattern recognition. pp. 3343--3352 (2019)

\bibitem{wang2021gdr}
Wang, G., Manhardt, F., Tombari, F., Ji, X.: Gdr-net: Geometry-guided direct
  regression network for monocular 6d object pose estimation. In: Proceedings
  of the IEEE/CVF Conference on Computer Vision and Pattern Recognition. pp.
  16611--16621 (2021)

\bibitem{NOCS}
Wang, H., Sridhar, S., Huang, J., Valentin, J., Song, S., Guibas, L.J.:
  Normalized object coordinate space for category-level 6d object pose and size
  estimation. In: Proceedings of the IEEE/CVF Conference on Computer Vision and
  Pattern Recognition. pp. 2642--2651 (2019)

\bibitem{wang2019frustum}
Wang, Z., Jia, K.: Frustum convnet: Sliding frustums to aggregate local
  point-wise features for amodal 3d object detection. In: 2019 IEEE/RSJ
  International Conference on Intelligent Robots and Systems (IROS). pp.
  1742--1749. IEEE (2019)

\bibitem{wohlhart2015learning}
Wohlhart, P., Lepetit, V.: Learning descriptors for object recognition and 3d
  pose estimation. In: Proceedings of the IEEE conference on computer vision
  and pattern recognition. pp. 3109--3118 (2015)

\bibitem{wu2020grasp}
Wu, C., Chen, J., Cao, Q., Zhang, J., Tai, Y., Sun, L., Jia, K.: Grasp proposal
  networks: An end-to-end solution for visual learning of robotic grasps.
  Advances in Neural Information Processing Systems  \textbf{33},  13174--13184
  (2020)

\bibitem{posecnn}
Xiang, Y., Schmidt, T., Narayanan, V., Fox, D.: Posecnn: A convolutional neural
  network for 6d object pose estimation in cluttered scenes. arXiv preprint
  arXiv:1711.00199  (2017)

\bibitem{pointfusion}
Xu, D., Anguelov, D., Jain, A.: Pointfusion: Deep sensor fusion for 3d bounding
  box estimation. In: Proceedings of the IEEE conference on computer vision and
  pattern recognition. pp. 244--253 (2018)

\bibitem{PRGCN}
Zhou, G., Wang, H., Chen, J., Huang, D.: Pr-gcn: A deep graph convolutional
  network with point refinement for 6d pose estimation. In: Proceedings of the
  IEEE/CVF International Conference on Computer Vision. pp. 2793--2802 (2021)

\end{thebibliography}
\end{document}